___________________________________________________________________

**Preliminary experiments on thermal emissivity adjustment for face images**

Marcos Faundez-Zanuy (1), Xavier Font Aragones (1), Jiri Mekyska (2)

(1) ESUP Tecnocampus (Pompeu Fabra University) Av. Ernest Lluch 32, 08302 Mataró, Spain.
(2) Department of Telecommunications, Brno University of technology, Czech Republic

faundez@tecnocampus.cat

**Abstract**

In this paper we summarize several applications based on thermal imaging. We emphasize the importance of emissivity adjustment for a proper temperature measurement. A new set of face images acquired at different emissivity values with steps of 0.01 is also presented and will be distributed for free for research purposes. Among the utilities, we can mention: a) the possibility to apply corrections once an image is acquired with a wrong emissivity value and it is not possible to acquire a new one; b) privacy protection in thermal images, which can be obtained with a low emissivity factor, which is still suitable for several applications, but hides the identity of a user; c) image processing for improving temperature detection in scenes containing objects of different emissivity.

1. **Introduction**

In the past we have used thermal images for a wide range of applications, including face biometric recognition, providing a new database freely available to the scientific community [1-3], hand morphology biometric recognition, including a hand image database distributed for free too [4-6], biomedical application for tuberculosis detection using tuberculine test and thermal imaging [7], and facial emotion recognition using thermal imaging in an induced emotion database [8]. We have also performed studies about the focusing of thermal images [9] and the fusion of different images containing objects at different focal distances [10-11].

In this paper we deal with a new research topic, which is the emissivity configuration in a thermal camera, and the possibility of several applications.

In order to obtain a correct temperature measurement, the emissivity adjustment of the thermal camera must be properly setup. Otherwise, wrong measurements are obtained. Table 1 represents some emissivity values [12-13]. The problem appears when several objects of interest with different emissivity are inside the scene. In this case, only the temperature of one of them can be directly estimated with a single thermal camera image.

In this paper we present a set of images acquired from a fixed scene varying the emissivity configuration of the camera. With a set of images acquired with different emissivity configuration, several research possibilities exist:

___

a) A combined image can be obtained with correct temperatures for all the objects inside the scene despites their different emissivity values. This can be achieved by means of image processing in a procedure similar to our previous work where we obtained a focused image containing objects at very different focal distance, being each object focused properly in a different image [10].

b) When dealing with human images privacy problems could appear [14]. However, in some applications such as activity surveillance of elder people inside their home, fall down detection, etc., a very precise detail is not necessary. A coarse description would be correct, for instance, to detect fall downs in the shower/ bathroom using thermal images. In this scenario, bad emissivity adjustment could be an alternative to hide the details of the face/body in order to protect the user's privacy. Especially in a situation where the user is typically naked.

c) When an image is acquired at wrong emissivity value and there is no possibility to acquire another one with proper settings, some kind of compensation/transformation can be done in order to correct the wrong adjustment effect. For this purpose, a set of images of the same object acquired at different emissivity values can help to establish the mathematical relation to perform this compensation.

In this paper we have worked with object and face images, and special effort is put on this last research possibility.

Table 1. Emissivity values for several materials

| Material | Emissivity |
|---|---|
| Aluminum foil | 0.03 |
| Aluminum, anodized | 0.9 |
| Asphalt | 0.88 |
| Brick | 0.90 |
| Concrete, rough | 0.91 |
| Copper, polished | 0.04 |
| Copper, oxidized | 0.87 |
| Glass, smooth (uncoated) | 0.95 |
| Human skin | 0.98 |
| Ice | 0.97 |
| Limestone | 0.92 |
| Marble (polished) | 0.89 to 0.92 |
| Paint (including white) | 0.9 |
| Paper, roofing or white | 0.88 to 0.86 |
| Plaster, rough | 0.89 |
| Silver, polished | 0.02 |
| Silver, oxidized | 0.04 |

___________________________________________________________________

| Material | Emissivity |
|---|---|
| Snow | 0.8 to 0.9 |
| Transition metal Disilicides (e.g. MoSi2 or WSi2) | 0.86 to 0.93 |
| Water, pure | 0.96 |

The paper is organized as follows: section 2 describes the experiments and databases acquired. Section 3 summarizes the main conclusions.

2.  **EXPERTIMENTS**

We have performed a first experiment with human faces, where we have acquired a static individual (his head is resting on the wall in order to make it easy to stay in a fix position for the whole set of acquisitions).

Figure 1 shows the relation between temperatures acquired in the middle of the eyebrows of a face image for different emissivity values. The images have been acquired with a TESTO 880 camera, which provides a sensibility of 100 mK and a resolution of 160x120 pixels without interpolation. This has been done by manual measurements using IRsoft from TESTO freely available for download from TESTO. Figure 2 shows the face image acquired at the correct emissivity (0.98) while Figure 3 shows a wrong set up for human skin (emissivity = 0.62).

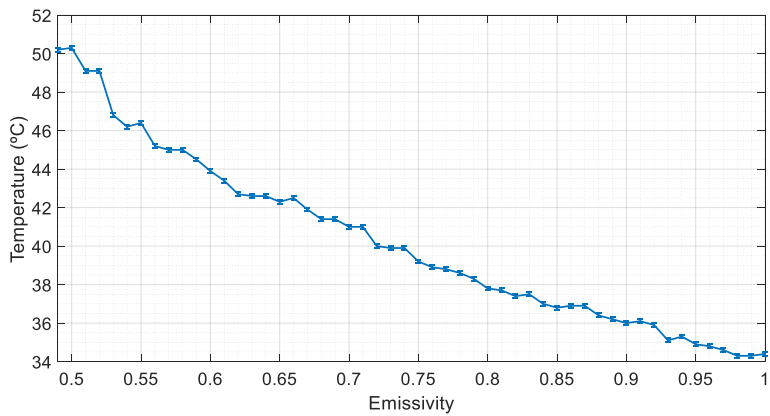

Figure 1. Relation between temperatures acquired in the middle of the eyebrows of a face image for different emissivity values.

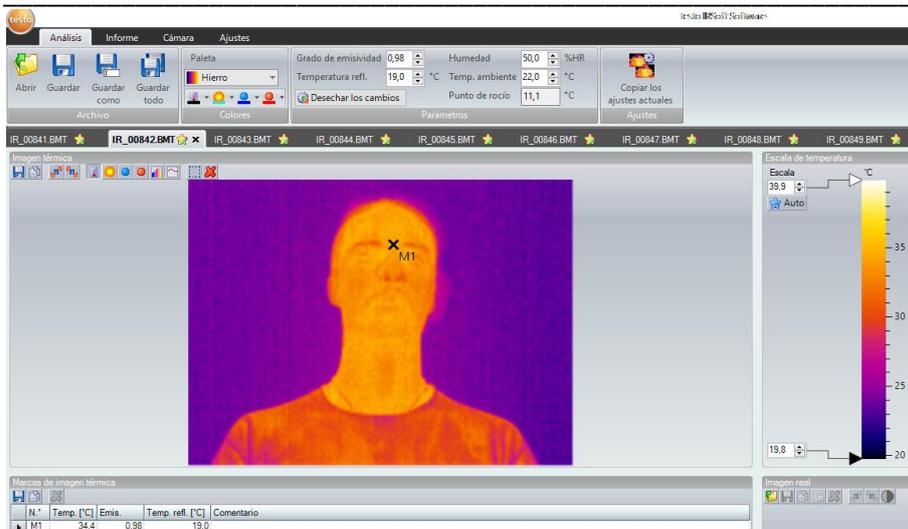

Figure 2. Temperature measurement at correct emissivity value.

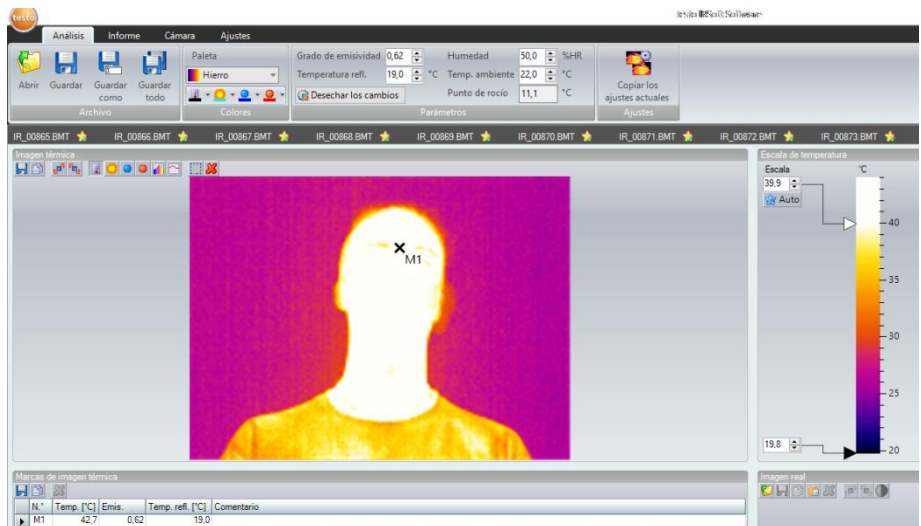

Figure 3. Temperature measurement at wrong emissivity value.

We also acquired a set of images of the same subject with a NEC H2640 camera. NEC H2640 provides a 640x480 pixels resolution without interpolation and resolution 0.06 °C or better (at 30 °C, 30 Hz). In this case, NEC provides proprietary software that requires a license key. This software permits an automatic analysis of the whole set of images. Figure 4 shows a screen shot of NEC software. On the bottom there is the automatic result of the analysis of the temperature in the ellipsoid in the centre of the eyebrows. From this analysis it is evident that fixing the emissivity to a value to 0.99 we always obtain the same result (almost flat plot for average, low and high temperature values in each acquisition). Thus, one important conclusion is that emissivity is important during visualization but not during acquisition.

Faundez-Zanuy, M., Font-Aragones, X., Mekyska, J. (2021). Preliminary Experiments on Thermal Emissivity Adjustment for Face Images. In: Esposito, A., Faundez-Zanuy, M., Morabito, F., Pasero, E. (eds) Progresses in Artificial Intelligence and Neural Systems. Smart Innovation, Systems and Technologies, vol 184. Springer, Singapore. https://doi.org/10.1007/978-981-15-5093-5_15

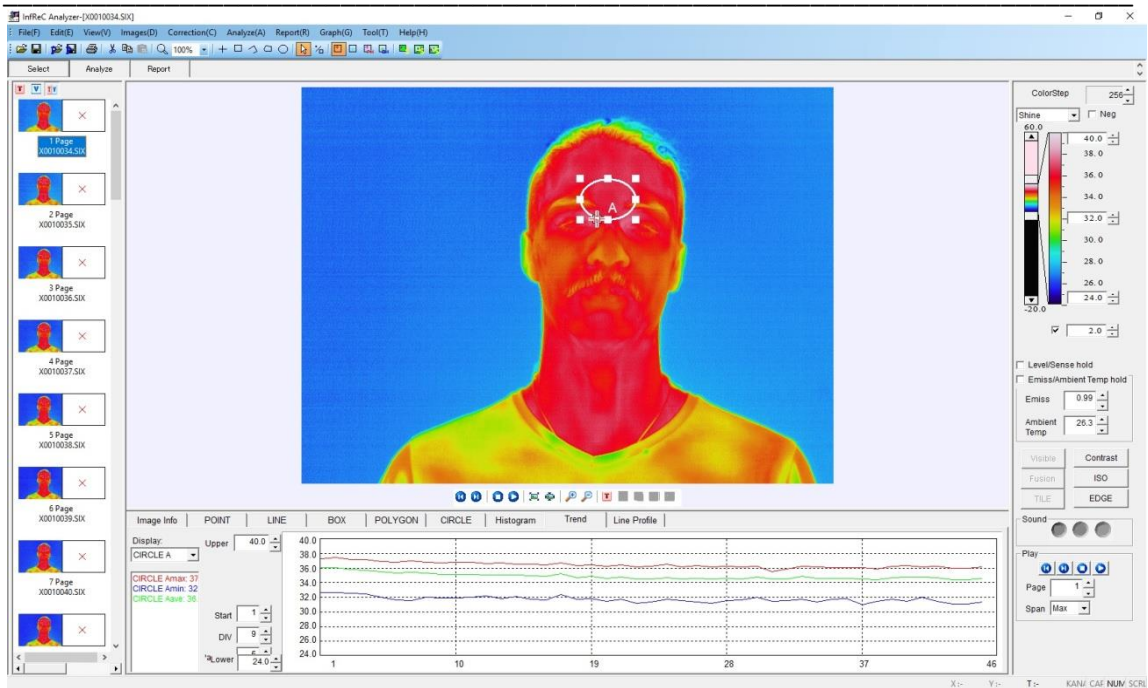

Figure 4. Temperature measurement with NEC software in a sequence of 46 snapshots acquired at different emissivity values.

In a second experiment we acquired a set of objects of different material, which has different emissivity (see table 1). Figure 5 shows the acquired scene. Different objects should be at same temperature (the room temperature) but they are visualized with different colour because of the different emissivity (wood, plastic, wax, chalk, metal). In this case the temperature analysis is performed inside the large square that covers almost the whole scene. This set of images can be used for future research analysis on measurements of objects with different emissivity at the same temperature. While a single image will be enough for the purpose, the existence of several consecutive acquisitions with small variations can be useful for algorithm development, which will not have to be based on a single image.

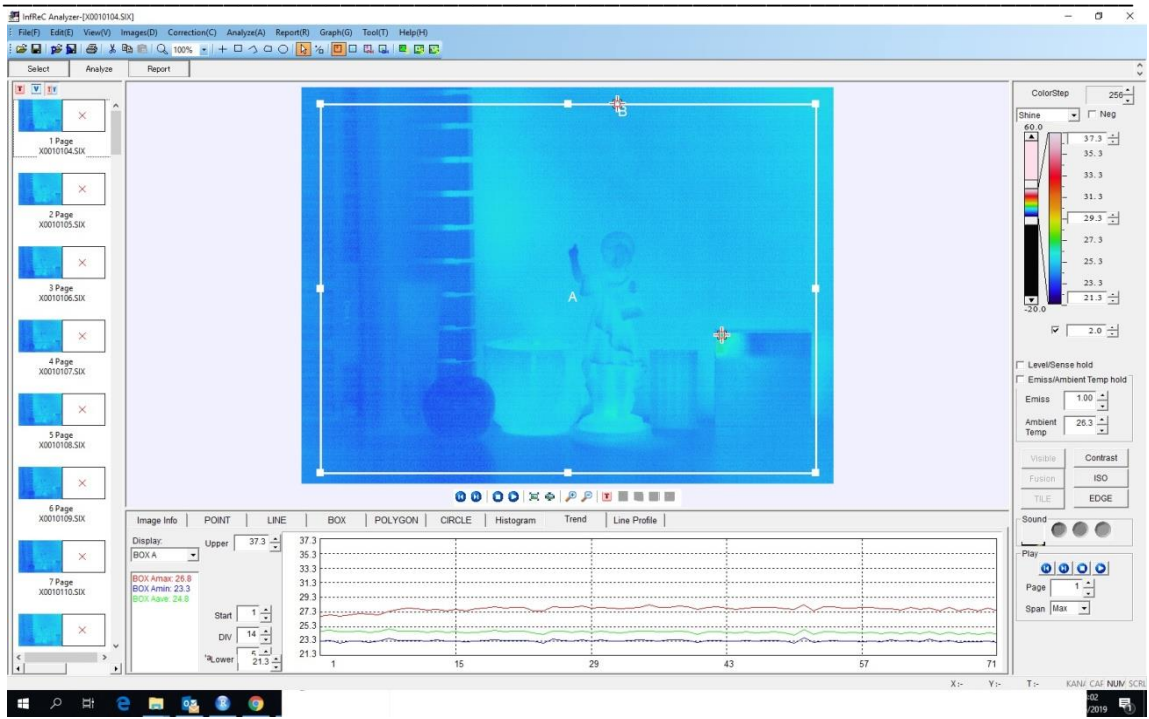

Figure 5. An image of a set of 71 images acquired of a fix scene modifying the emissivity configuration in each image.

3. **Conclusions**

In this paper we dealt with two thermal cameras and their own infrared image processing software. These cameras have been a TESTO 880 and NEC H2640. With these cameras we acquired three sets of images where each image has a different emissivity configuration. Based on these sequences we can establish:

a) The emissivity value has influence during visualization of the image but not during acquisition. During the acquisition the image file stores the values provided by the sensor and analog to digital converter. During visualization these raw values are converted into pseudocolors using the emissivity information provided by the analysis software, which can be the same fix during acquisition or different.
b) User privacy can be protected in some way visualizing the image with a wrong emissivity configuration. However, privacy can be compromised once the adjustment returns to correct value. In this context privacy is considered as the possibility of indentifying the user inside the image and/or the possibility "to see" the shape and lengths of parts of his body.
c) Measurement of different objects, each one with different emissivity, is a problem that cannot be addressed with current thermal cameras. This is due to the fact that emissivity is adjusted for the whole image. Some algorithm and software has to be developed for segmenting objects, assigning emissivities to each of them, and measuring proper values. This remains an open research issue.

Faundez-Zanuy, M., Font-Aragones, X., Mekyska, J. (2021). Preliminary Experiments on Thermal Emissivity Adjustment for Face Images. In: Esposito, A., Faundez-Zanuy, M., Morabito, F., Pasero, E. (eds) Progresses in Artificial Intelligence and Neural Systems. Smart Innovation, Systems and Technologies, vol 184. Springer, Singapore. https://doi.org/10.1007/978-981-15-5093-5_15

______________________________________________________________________

## Acknowledgement

This work has been supported by FEDER and MEC, TEC2016-77791-C4-2-R, PID2019-109099RB-C41 and LO1401.

## 4. References

[1] Virginia Espinosa, Marcos Faundez-Zanuy and Jiri Mekyska "Beyond cognitive signals" Cognitive Computation. Springer. Vol. 3 pp.374–381, June 2011 https://doi.org/10.1007/s12559-010-9035-6

[2] Virginia Espinosa, Marcos Faundez-Zanuy, Jiri Mekyskya and Enric Monte, "A criterion for analysis of different sensor combinations with an application to face biometrics" Cognitive Computation. Volume 2, Issue 3 (2010), Page 135-141. September 2010 https://doi.org/10.1007/s12559-010-9060-5

[3] Jiri Mekyska, Virginia Espinosa-Duró and Marcos Faundez-Zanuy, "Face segmentation: a comparison between visible and thermal images" IEEE 44th International Carnahan Conference on Security Technology ICCST 2010, San José, USA. 5-8 octubre 2010 doi: 10.1109/CCST.2010.5678709

[4] Marcos Faundez-Zanuy, Jiri Mekyska, Xavier Font-Aragonès "A new hand image database simultaneously acquired in visible, near-infrared and thermal spectrums". Cognitive computation. Vol. 6 Number 2, Cogn Comput (2014) 6:230-240 https://doi.org/10.1007/s12559-013-9230-3

[5] Xavier Font-Aragones, Marcos Faundez-Zanuy, Jiri Mekyska "Thermal hand image segmentation for biometric recognition". IEEE Aerospace and Electronic Systems Magazine. June 2013, vol. 28, num. 6, p. 4-14 doi: 10.1109/MAES.2013.6533739

[6] Jiri Mekyska, Xavier Font, Marcos Faundez-Zanuy, Rubén Hernández-Mingorance, Aythami Morales, Miguel Ángel Ferrer-Ballester, "Thermal hand image segmentation for biometric recognition"·pp.26-30, 45th IEEE Carnahan Conference on Security Technology ICCST'2011, 18-21 october 2011, Mataró. doi: 10.1109/CCST.2011.6095877

[7] Jose Antonio Fiz, Manuel Lozano, Enric Monte-Moreno, Adela González-Martínez, Marcos Faundez-Zanuy, Caroline Becker, Laura Rodriguez-Pons, Juan Ruiz Manzano "Tuberculine reaction measured by infrared thermography" Computer Methods and Programs in Biomedicine, Volume 122, Issue 2, November 2015, Pages 199-206. https://doi.org/10.1016/j.cmpb.2015.08.009

[8] Anna Esposito, Vincenzo Capuano, Jiri Mekyska, Marcos Faundez-Zanuy "A naturalistic database of thermal emotional facial expressions and effects of induced emotions on memory" Pages: 158-173 Proceedings of the 2011 international conference on Cognitive Behavioural Systems. LNCS Springer-Verlag Berlin, Heidelberg ©2012 ISBN: 978-3-642-34583-8 Dresden, February 2011 https://doi.org/10.1007/978-3-642-34584-5_12

Faundez-Zanuy, M., Font-Aragones, X., Mekyska, J. (2021). Preliminary Experiments on Thermal Emissivity Adjustment for Face Images. In: Esposito, A., Faundez-Zanuy, M., Morabito, F., Pasero, E. (eds) Progresses in Artificial Intelligence and Neural Systems. Smart Innovation, Systems and Technologies, vol 184. Springer, Singapore. https://doi.org/10.1007/978-981-15-5093-5_15

___________________________________________________________________

[9] Marcos Faundez-Zanuy, Jiri Mekyska and Virginia Espinosa "On the focusing of thermal images" Pattern Recognition letters. Elsevier. Vol. 32 (2011) pp. 1548–1557, August 2011. https://doi.org/10.1016/j.patrec.2011.04.022

[10] Radek Benes, Pavel Dvorak, Marcos Faundez-Zanuy, Virginia Espinosa-Duro, Jiri Mekyska "Multi-focus thermal image fusión", Pattern Recognition Letters 34 Issue 5, 1 April (2013) 536–544. https://doi.org/10.1016/j.patrec.2012.11.011

[11] Virginia Espinosa-Duró, Marcos Faundez-Zanuy, Jiri Mekyska "Contribution of the Temperature of the Objects to the Problem of Thermal Imaging Focusing" 46 ICCST'2012, pp.363-366. Boston, USA IEEE Catalog Number: CFP12ICR-USB ISBN: 978-1-4673-2449-6. doi: 10.1109/CCST.2012.6393586

[12] Brewster, M. Quinn (1992). Thermal Radiative Transfer and Properties. John Wiley & Sons. p. 56. ISBN 9780471539827.

[13] 2009 ASHRAE Handbook: Fundamentals - IP Edition. Atlanta: American Society of Heating, Refrigerating and Air-Conditioning Engineers. 2009. ISBN 978-1-933742-56-4.

[14] Marcos Faundez-Zanuy "Privacy issues on biometric systems". IEEE Aerospace and Electronic Systems Magazine. Vol.20 nº 2, pp13-15, ISSN: 0885-8985. February 2005. DOI: 10.1109/MAES.2005.9740719